# Long-Term Progress and Behavior Complexification in Competitive Co-Evolution

Luca Simione and Stefano Nolfi
Institute of Cognitive Sciences and Technologies
National Research Council (CNR-ISTC)
Rome, Italy
luca.simione@istc.cnr.it, stefano.nolfi@istc.cnr.it

**Abstract**
*The possibility to use competitive evolutionary algorithms to generate long-term progress is normally prevented by the convergence on limit cycle dynamics in which the evolving agents keep progressing against their current competitors by periodically rediscovering solutions adopted previously over and over again. This leads to local but not to global progress, i.e. progress against all possible competitors. We propose a new competitive algorithm that produces long-term global progress by identifying and by filtering out opportunistic variations, i.e. variations leading to progress against current competitors and retrogression against other competitors. The efficacy of the method is validated on the co-evolution of predator and prey robots, a classic problem that has been used in other related researches. The accumulation of global progress over many generations leads to effective solutions that involve the production of rather articulated behaviors. The complexity of the behavior displayed by the evolving robots increases across generations although progresses in performance are not always accompanied by behavior complexification.*

## 1. Introduction

Competitive co-evolution, i.e. the evolution of populations with coupled fitness, present important potential advantages.

First, the co-evolution of competing species such as predator and prey might favor the synthesis of evolutionary innovations. Indeed, an adaptation in one lineage (e.g. predators) may change the selection pressure on another lineage (e.g. prey), giving rise to a counter-adaptation. If this occurs reciprocally, "an unstable runaway escalation of 'arm races' may result." (Dawkins and Krebs, 1979, p. 55). In other words, adaptations on one side call for counter adaptations on the other side, and the counter adaptations call for more counter adaptations and so on, thus producing an escalation process.

Secondly, competitive co-evolution can potentially produce a self-regulating incremental process in which the complexity of the adaptive problem is tuned to the ability of the evolving agents and increases across generations. In fact, in competitive scenarios the complexity of the problem depends primarily on the efficacy of the competitors that become better and better as the skills of the evolving agents increase. Exposing agents to progressively harder conditions and to conditions that match their skill level can facilitate the discovery of progressively better strategies (Rosin and Belew, 1997).

Thirdly, competitive co-evolution methods constitute a natural choice for problems, such as game-playing, in which identifying an absolute quality measure is difficult or not possible (Chong. Tino, Ku and Yao, 2012).

Unfortunately, the occurrence of arm races leading to global progress for a prolonged period of time is only one of the possible outcomes of a competitive co-evolutionary dynamics. "One side may drive the other to extinction; one side might reach a definable optimum, thereby preventing the other side from reaching its optimum; both sides might reach a mutual local optimum; or the race may persist in a theoretically endless limit cycle" (Dawkins and Krebs, 1979, pp.70). Moreover, prolonged progress against competitors does not necessarily imply global progress (Miconi, 2009), i.e. the development of strategies that are more and more effective not only against the current competitors (local progress) but also against ancient competitors (historical progress) and against new competitors (global progress).

Indeed, the competitive co-evolutionary experiments carried out to date have not shown evidences of long-term global progress (Miconi, 2008; Nolfi, 2012). The most common outcome is a limit cycle dynamic in which qualitatively similar strategies are abandoned and rediscovered over and over again (Cliff and Miller, 1996; Nolfi and Floreano, 1998; Watson and Pollack, 2001). To understand the nature of this dynamics, suppose that at a certain evolutionary stage population 1 adopts the strategy A that is effective against the strategy B currently adopted by the competing population 2. Imagine now that there is a strategy C (genetically similar to B) that is more effective against the strategy A. Population 2 will abandon the strategy B in favor of C. Imagine now that there is a strategy D (genetically similar to A) that is effective against the strategy C. Population 1 will abandon the strategy A in favor of D. Finally imagine that the previously strategy B is effective against strategy D. Population 2 will abandon C and will return to B. At this point, also population 1 will also return to A (because, as explained above, A is effective against B). This implies that the two populations return to their initial strategies A and B and will then keep re-adopting C and D and A and B again and again. Cycling dynamics of this type were found in natural evolution, for example in in population of side-blotched lizards (*Uta stansburiana*) by Sinervo and Lively (1996; for a discussion of less regular form of cycling see Cartlidge and Bullock, 2004).

In this paper we demonstrate for the first time how co-evolutionary experiments involving predator and prey robots can lead to long term global progress. This is achieved by using an anti-opportunistic algorithm that periodically divides the populations in a training and a validation sets and uses the validation set to identify and to filter out the variations leading to local progress only, i.e. to progress against the training opponents accompanied by retrogression against the validation opponents. Moreover, we demonstrate how the rate of global progress can be increased by exposing evolving robots to well-differentiated competitors and by preserving individuals displaying good performance against hard to handle competitors. Finally, we show how the behaviors of the robots evolved with our algorithm become progressively more complex across generations, although progress in performance are not always accompanied by behavior complexification.

In section 2 we describe the method proposed and the relation with the state of the art. In section 3 we report our results. In section 4 we analyze the complexity of the behavior displayed by evolving agents across generations. In section 5 we report additional experiments performed in a more complex environment. Finally, in section 6, we draw our conclusions.

## 2. Method and relation with the state-of-the-art

Competitive co-evolution involves two evolving populations with coupled fitness, e.g. predators and prey or hosts and parasites. Alternatively, it can involve a single population of competing individuals, e.g. fighting agents or game players in symmetrical games. The fitness of individuals is computed during multiple evaluation episodes in which the individuals interact with different opponents.

In this paper we focus on the maximization of the expected utility, i.e. the performance against all possible opponents. Other authors investigated the usage of competitive co-evolution for the synthesis of Nash equilibrium solutions (Ficici, 2004; Ficici & Pollack, 2003; Wiegand, Liles and De Jong, 2002) and Pareto-optimal solutions (De Jong, 2004). The maximization of the expected utility has a high practical relevance since it is not constrained by the problems and limits of alternative approaches. More specifically, by the fact that Nash equilibrium solutions do not necessarily correspond to high performing solutions. Moreover, the fact that the optimization of Pareto optimal solutions can be applied to deterministic problems only and often require the optimization of a number of objectives that is too large for a practical search algorithm (De Jong, 2005).

Previous attempts to maximize the expected utility focused on the utilization of archives of ancient opponents or on the usage of randomly generated opponents. The idea of preserving individuals from previous generations and of evaluating agents against current and ancient opponents has been introduced by Rosin and Belew (1997). Their method preserves the best individuals of each generation in a "hall-of-fame" archive and evaluates evolving agents against all the opponents contained in the archive.

These technique guarantees historical progress but not global progress, i.e. the development of solutions that are better and better against all possible competitors. Indeed, as reported in Nolfi and Floreano (1998), predator and prey robots evolved against all the members of the all-of-fame archive display increasing better performance against ancient robots included in the archive but produce solutions that are less effective than those produced with a vanilla algorithm that does not use the archive. This can be explained by considering that evaluating agents against an increasing number of ancient competitors facilitates the premature converge toward local optima (Nolfi and Floreano, 1998). More generally, it leads to the generation of a limited number of qualitatively different opponents, to solutions that are overfitted to the opponents contained in the archive, and to solutions that generalize poorly to other opponents (Miconi, 2009).

A possible better algorithm is the Maxsolve introduced by De Jong (2005) which maintains an archive containing a limited number of agents. In each iteration, the algorithm receives a new set of agents and a new set of opponents that might or might not be included in the archive. The selection is performed by discarding solutions identical to those already included in the archive and by replacing the agents of the archive with the worst performance with the new agents with the best performance, providing that the latter outperform the former.

Promising results were collected by using a variation of the Maxsolve method by Samothrakis, Lucas, and Runarsson (2013) in the context of the co-evolution of Othello game players. In their model the archive has a limited size, the evolving agents are evaluated against all the opponents contained in the archive, and the archive is updated only when a new agent outperform all the opponents included in the archive (in that case, the new agent is used to replace the oldest member of the archive). As remarked by the authors, however, this technique is only applicable to problems in which the outcome of the evaluation episodes is deterministic, e.g. in perfect games (Samothrakis, Lucas, and Runarsson, 2013).

An alternative approach consists in using randomly generated opponents (Chong, Tino, and Yao, 2008; Chong, Tino, Ku, and Yao, 2012). The potential advantage of this technique is the direct promotion of global progress due to the fact that the agents are evaluated against always new opponents. The disadvantage is that the efficacy of the opponents does not increase across generations. Consequently, the method does not permit to develop agents capable of defeating strong opponents. Indeed, the results obtained through this method by Samothrakis, Lucas, and Runarsson (2013) were much poorer with respect to the performance obtained by using a variation of the Maxsolve algorithm illustrated above.

In the context of reinforcement learning, the utilization of competitive multi-agent scenarios recently produced remarkable results for the training of game players (Silver et al., 2016; Heinrich and Silver, 2016) and simulated robots (Bansal et al., 2018).

**2.1 The generalist algorithm**

We propose a new algorithm that promotes the evolution of global progress by filtering out opportunistic solutions. This is realized by periodically: (i) evolving a subset of a population (agents) against a subset of the competing population (opponents), (ii) periodically evaluating the evolved agents against all opponents, and (iii) filtering out opportunistic agents that perform well against the selected opponents but poorly against the other opponents.

More specifically the evolutionary process (see also the pseudo-code below) is organized in a series of phases in which a subset $n$ of the first population is evolved against a subset $n$ of the second population for a certain number of generations (where $N$ is the size of each of the two populations and $n < N$) and vice versa.

The former and the latter subsets are referred as the agents and the opponents, respectively. The remaining opponents are referred as the validation opponents.

At the beginning of each phase, the two subsets are chosen (see the explanation below). Then for a certain number of generations, the agents are evolved against the opponents. Each agent is evaluated for *n* episodes against each opponent. At the end of each phase, the population of *N* agents is replaced with the best *N* individuals selected among the N original agents and the *n* evolved agents. The best agents are identified by ranking the *N+n* individuals on the basis of the average performance obtained against all opponents (i.e. against the opponents and the validation opponents).

The filtering of opportunistic individuals occurs in this last phase, since agents with enhanced performance with respect to the opponents and reduced performance against validation opponents are discarded. Similar techniques are used in the field of machine learning to avoid overfitting, i.e. to avoid the retention of variations that produce improvements with respect to the training set and retrogressions with respect to the validation set (Searle, 1995; Srivastava at al., 2014). The method proposed can also be considered an efficient heuristic for sampling the "maximally informative test set" discussed by Bucci and Pollack (2002).

Using *n<N* is necessary to leave *N-n* individuals for the validation set. In our experiments we set *N* to 80 and *n* to 10. Preliminary results suggest that qualitatively similar results can be obtained by setting *N* to 20. We did not collected results by varying *n*. We use a full pairwise competition in which each agent is evaluated against each opponent. Best versus best method (Sims, 1994) or tournament selection (Angeline and Pollack, 1993) allow to reduce the number of evaluations but increase the risks of overspecialization.

---

**Generalist algorithm (Standard condition)**

1. **Start**
2. *N* = 80, *n* =10, *nphases*=1500, ngenerations=100, current_generation=0
3. **Initialize** the genome of pop1 and pop2 populations
4. use predator as evolving agents and prey as opponents
5. **for** phase in [0, *nphases*]
6.     **select** *n* opponents through the clustering method (see text)
7.     **select** *n* agents (i.e. the agents with the highest fitness against each selected opponent)
8.     **for** generation in [0, ngenerations]
9.         **create** *n* offspring (i.e. *n* mutated copies of the *n* selected evolving individuals)
10.         **evaluate** *n* offspring for n episodes against each opponent
11.         **replace** each parent with each offspring if the fitness of the latter is >= of the fitness of the former
12.         **current_generation** += 1
13.         **if** ((current_generation % 500) == 0) invert agents and opponents
14.     **rank** *N+n* agents on the basis of their weighted performance against N opponents (see text)
15.     **replace** the evolving population with the top N agents selected among the N+n agents
16. **End**

---

In the case of the experiments reported in this paper, agents are evolved through a (1+1) evolutionary strategy (Rechenberg, 1973). Each agent is allowed to generate an offspring, i.e. a copy with mutations. Mutations are realized by replacing 2% of the genes with floating-point values selected randomly with a uniform distribution in the range [-5.0, 5.0]. Each parent is replaced by its corresponding offspring if the fitness of the latter us greater or equal to the fitness of the former. However, the generalist method is a meta heuristic that can be used in combination with any black box algorithm (e.g. CMA-ES [Hansen and Ostermeier, 2001]).

The *n* opponents are selected by clustering all opponents into *n* groups on the basis of the similarity of the fitness achieved by them against all agents, and by choosing the fittest opponent of each group. The *n* opponents are then chosen by selecting the best opponent of each group. Similarity is calculated on the basis of the Euclidean distance between the *N* vectors of *N* values that encode the performance of each opponent against each agent. The clustering is realized in 3 steps: first, each opponent is clustered with the most similar opponent so to form *N*/2 groups, then each group is clustered with the closest other group so to form *N*/4 and finally *N*/8 groups. This clustering method permits to select high performing opponents among well-differentiated group formed by the same number of opponents. Notice that opponents are differentiated with respect to the strategies that are effective against them and not with respect to the behavior that they produce.

The *n* agents are selected by choosing the agents that achieved the highest fitness against each of the *n* selected opponents. This allows to select agents displaying high performance against each opponent although not necessarily against all opponents.

The ranking of *N*+*n* agents at the end of each phase is performed by multiplying the fitness achieved against each of the *N* opponents for the average fitness obtained by the opponent. The utilization of this weighted ranking techniques (Rosin and Belew, 1997) permits to preserve agents that perform well against high-performing opponents.

The utilization of alternated evolutionary phases in which a population evolves against a fixed set of opponents and vice versa allows to avoid the need to re-evaluate the performance of the original agents against the opponents. Moreover, it allows to reduce the speed with which the social environment of the agents varies. As shown by Milano et al. (2017a, 2017b), agents evolved in environments that vary at a moderate rate (i.e. every N generations) achieve better performance than agents evolved in environments that vary every generation (Milano et al. 2017a, 2017b).

To verify the importance of selecting well-differentiated opponents and the importance of preserving agents performing well against high-performing opponents we also carried out experiments with a variation of this algorithm, indicated as Simplified algorithm, in which: (i) the opponents are selected by choosing the *n* individuals with the highest average fitness against all agents, and (ii) *N*+*n* agents are ranked directly on the basis of their average fitness, rather than by using the weighted ranking described above.

Finally, to verify the importance of filtering out opportunistic individuals we also carried out experiments with a second control algorithm, in which all agents are evaluated against all opponents (i.e. in which *n*=*N*). This method corresponds to a Vanilla co-evolutionary algorithm that does not include a method for filtering out opportunistic agents.

## 2.2 The predators and prey problem

We decided to test our algorithm on the co-evolution of predator and prey robots since it is a challenging problem both for natural organisms and autonomous robots (Miller and Cliff, 1994) and since it has widely used as a test-bed for competitive co-evolutionary algorithm (Miller and Cliff, 1994, Haynes and Sen, 1996; Floreano and Nolfi, 1997a; Floreano and Nolfi, 1997b; Nolfi and Floreano, 1998; Floreano, Nolfi and Mondada, 1998; Stanley and Mikkulainen, 2002; Buason and Ziemke, 2003; Buason, Bergfeldt and Ziemke, 2005; Janssen at al., 2016). Indeed, the need to face highly dynamic, largely unpredictable, and hostile environments requires the development of reactive, robust, and reliable solutions. Moreover, mastering predators and prey competition requires the ability to display several integrated behavioral and cognitive capabilities such as avoiding fixed and moving obstacles, optimizing the motion trajectory with respect to multiple constraints, integrating sensory information over time, anticipating the behavior of the opponent, managing appropriately available energy resources, disorienting the opponent and copying with protean defense behaviors (Humphries and Driver, 1970), adapting on the fly to the behavior of the current opponent.

The robots are constituted by simulated MarxBot (Bonani et al., 2010) provided with neural network controllers. The connection strength of the robots' neural network that determine the robots' behavior are encoded in artificial genotypes and evolved. Predators are evolved for the ability to capture prey (i.e. to reach and physically touch the prey) as fast as possible and prey are evolved for the ability to avoid being captured as long as possible. The fitness of the prey and of the predators corresponds to fraction of time required by the predator to capture the prey and to the inverse of the fraction of time required by the predator to capture the prey, respectively. A detailed description of the characteristics of the robots and of the robots' neural controllers is provided in the Appendix.

The experiments reported in this paper can be replicated by downloading and installing FARSA, which is available from https://sourceforge.net/projects/farsa/, and by downloading and installing the experimental plugin available from http://laral.istc.cnr.it/res/predprey2019/ Simione_Nolfi_2019_plugin.zip. Videos of the evolved behaviors and additional supplementary material are available from http://laral.istc.cnr.it/res/predprey2019/

## 2.3 Measuring progress

In experiments involving a single population of agents evolving in non-varying environments progress corresponds to increase of fitness over generations. Instead, in competitive co-evolutionary experiments in which the social environment varies, the fitness cannot be used to infer directly the efficacy of individuals. This is due to the fact that the fitness of individuals does not only depend on the characteristics of the agents but also on the characteristics of the competitors that vary throughout generations.

This implies that, as pointed out by Van Valen (1973), evolving populations should adapt at the same rate of their competitors to preserve the same fitness level. Indeed, the term "Red Queen" introduced by Van Valen (1973) to characterize the dynamics of multiple evolving populations is derived from the statement that the Red Queen makes to Alice in Lewis Carroll's "Through the Looking-Glass" novel: "Now, here, you see, it takes all the running you can do, to keep in the same place". Phases during which the fitness of the two populations do not vary can correspond either to a stagnation phase, in which the two populations do not improve, or to a progress phase in which both populations improve and in which the advantage gained by the retention of adaptive variations in one population is compensated by the retention of adaptive variations in the opponent population. Moreover, phases during which the fitness of one population increases can correspond either to an improvement of that population or to a regression of the opponent population.

Consequently, measuring progress in a competitive co-evolutionary scenario requires the utilization of more elaborated measures such CIAO and master tournament. CIAO plots (Cliff and Miller, 1995, 2006) are obtained by post-evaluating individuals against competitors of previous generations and are consequently useful to measure historical progress. Master tournament analysis (Nolfi and Floreano, 1998) are obtained by post-evaluating individuals against competitors of previous and future generations and are consequently useful to measure both historical and global progress. The relative efficacy of alternative experimental conditions with respect to global progress can be evaluated by testing the individuals evolved in a first experimental condition with the competitors evolved in a second experimental condition (Nolfi and Floreano, 1988; Miconi, 2009). Finally, global progress can also be measured by post-evaluating the individuals of successive generations of an evolutionary experiment against the opponents evolved in other replications of the experiment (Miconi, 2009). In other words, the opponents obtained in another replication can be used as a validation set.

The duration of progress indicates the number of generations during which the evolving agents keep increasing their performance against competitors of previous generations (historical progress) and/or previous and future generations (global progress). The rate of progress corresponds to the ratio between the amount of performance increase, measured by post-evaluating the individuals against competitors of previous generations, and the number of generations during which such increase was produced.

## 3. Results

Here we report the results obtained by evolving the predator and prey robots for 150,000 generations with the generalist algorithm (Standard) and with the two control algorithms (Simplified and Vanilla). Five replication experiments for each condition were run. The predator and prey population size ($N$) was set to 80. The number of agents and opponents ($n$) was set to 10. Evolving phases last 100 generations.

As expected, the fitness of the predator and of the prey increases and decreases respectively during the phases in which predators evolve and decreases and increases respectively during the phases in which prey evolve. For sake of clarity, Fig. 1 reports data for the first and last 5000 generations (upper and lower panel). Sudden variations occurring every 100 generations are caused by the selection of new agents and opponents (see Section 2.1). Notice how the evolving robots improve their performance against their current competitors through the entire evolutionary process. This indicates that the co-evolutionary dynamics never converges on a stable state. The evolving individuals always maintain an ability to progress against their current competitors. This result is in line with the data collected in many previous co-evolutionary experiments (Cliff and Miller, 1996; Nolfi and Floreano, 1998; Buason, Bergfeldt and Ziemke, 2005; Stanley and Mikkulainen, 2002; Janssen at al., 2016) in which co-evolving individuals keep changing until the end of the evolutionary process by never converging on a stable state.

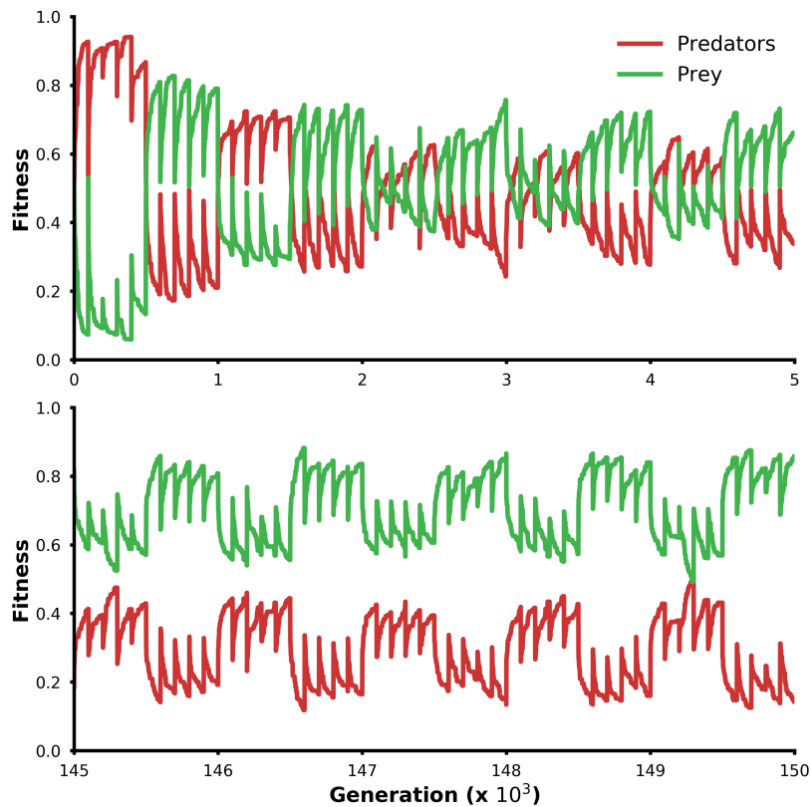

Fig. 1. Average fitness of the evolving robots ($n$ agents) throughout generations. The red and green curves correspond to predator and prey robots, respectively. Data plotted every generation. For sake of clarity we displayed only the first and the last 5000 generations of the first replication of the experiment (top and last panel, respectively). The other replications produced qualitatively similar results.

Fig. 2 displays the average fitness of the two populations of 80 individuals evaluated against each other throughout generations. As can be seen, the relative fitness of prey increased over generations whereas predators' fitness decreased. This implies that, overall, the problem of capturing prey is more difficult than the problem of escaping predators in our experimental condition. Prey, however, never managed to fully defeat predators. This implies that both species have the possibility to improve during the entire co-evolutionary process. Notice also that for the reasons discussed in section 2.3, the fact that the relative fitness of the two populations was relatively stable throughout generations does not imply that the behavior of the agents remained stable. As we will illustrate below, predator and prey became more and more effective throughout generations. The fitness remains relatively stable because progress of one species are generally compensated by progress of the other species.

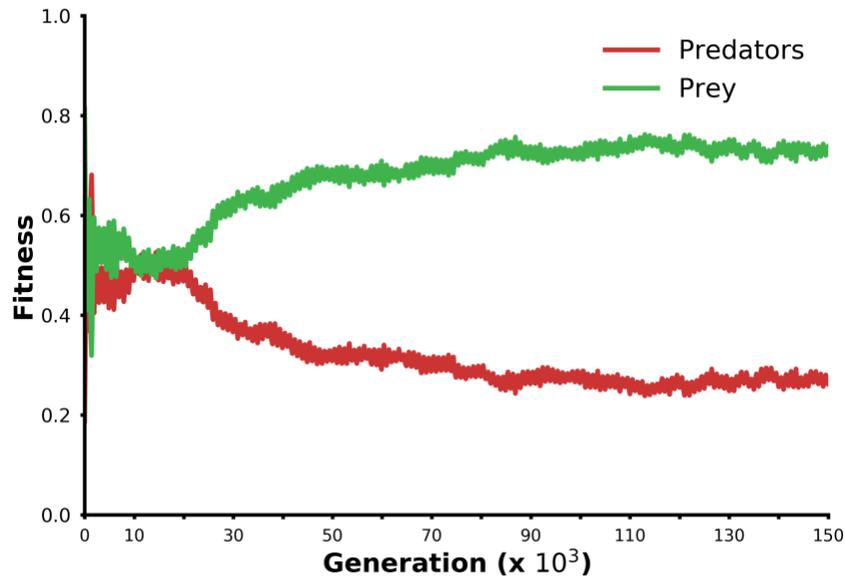

Fig. 2. Average fitness of predator and prey populations (N individuals) evaluated against each other throughout generations. The red and green curves correspond to predators and prey, respectively. Data plotted every 100 generations. Data of the first replication of the Standard experimental condition. The other replications produced qualitatively similar results.

To verify whether co-evolution leads to historical and global progress, we post-evaluated the two populations every 10,000 generations against competitors of previous and future generations, selected every 10,000 generations. The results indicate the occurrence of both historical and global progress (see Fig. 3). Indeed, robots at any particular generation generally performed better against ancient competitors than against current competitors. For example, predators of generation 150,000 achieved a fitness of about 0.25 against contemporary competitors and a fitness of about 0.6 against competitors of generation 10,000. Prey of generations 150,000 achieve a fitness of about 0.75 against contemporary competitors and a fitness of about 0.9 against competitors of generation 10,000. To appreciate the implications of these differences, it is important to point out that relatively small improvements in term of fitness corresponded to large improvements in term of behavioral capabilities (see below and the additional online materials). Preliminary results of these experiments were reported in Simione and Nolfi (2017).

Moreover, robots at any particular generation generally performed better against opponents of future generations than robots of previous generations. This implies that the strategies acquired by predator and prey of succeeding generations are not only more and more effective against previous opponents but also more and more effective in general, i.e. are more and more capable of defeating opponents, including opponents that they never encountered.

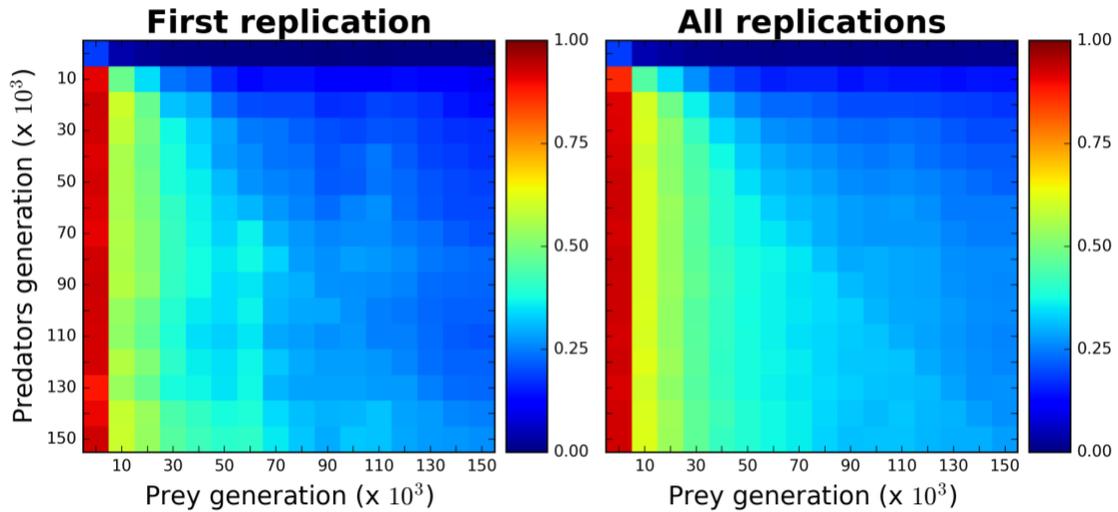

Fig. 3. Performance of predators and prey of every 10,000 generations post-evaluated against competitors of previous and following generations (master tournament). For each comparison, we indicate the performance of the predator. The performance of the prey corresponds to 1.0 minus the performance of the predator. The left and right figures display the results of the first replication and the average result of 5 replications, respectively.

**Table 1. Historical progress measured as the most recent generation of ancient opponents against which evolving agents obtained significantly better performance than against current opponents.**

| Agents generation | Standard | | Simplified | | Vanilla | |
| --- | --- | --- | --- | --- | --- | --- |
| | Predators | Prey | Predators | Prey | Predators | Prey |
| 1 | 0 | 0 | 0 | 0 | 0 | 0 |
| 2 | 1 | 1 | 1 | 1 | 0 | 0 |
| 3 | 2 | 1 | 1 | 1 | 0 | 0 |
| 4 | 2 | 2 | 2 | 2 | 0 | 0 |
| 5 | 3 | 2 | 3 | 2 | 0 | 0 |
| 6 | 4 | 3 | 4 | 2 | 0 | 0 |
| 7 | 4 | 3 | 4 | 2 | 0 | 0 |
| 8 | 5 | 4 | 4 | 3 | 0 | 0 |
| 9 | 5 | 5 | 4 | 3 | 0 | 0 |
| 10 | 5 | 4 | 4 | 5 | 0 | 0 |
| 11 | 5 | 4 | 5 | 5 | 0 | 0 |
| 12 | 6 | 4 | 5 | 5 | 0 | 0 |
| 13 | 7 | 4 | 5 | 5 | 0 | 0 |
| 14 | 7 | 7 | 6 | 6 | 0 | 0 |
| 15 | 7 | 7 | 6 | 6 | 0 | 0 |

*Note. Results relatively to the three evolving algorithms are separately reported in the left (Standard), middle (Simplified), and right (Vanilla) columns. Generations are expressed in x10$_3$, i.e. 1=10,000, 2=20,000 and so on. Generation 0 refers to the very first generation of randomly generated individuals. Current opponents refer to the*

*opponent of the same generation. Comparison between performances were conducted by using a one-tailed t-test, with N=5 replications.*

For what concerns historical progress, Table 1 shows the most recent generation of ancient opponents that achieved significant lower performance than current opponents against each tested generation of agents (1-tailed t-tests, N=5 replications). As shown in the table, in the case of the standard algorithm, historical progress continued until generation 130,000 for predators and until generation 140,000 for prey. Indeed, predators of generation 130,000 produced significantly better performance against prey of generation 70,000 than against prey of generation 130,000. Moreover, prey of generation 140,000 achieved significantly better performance against predator of generation 70,000 than against contemporary prey. In the case of the Simplified algorithm, historical progress continued up to generation 140,000 for both predators and prey. Instead, the Vanilla algorithm did not show any evidence of historical progress. In fact, performance against ancient competitors were not significantly better than performance against recent competitors, independently from how ancient the former competitors were.

For what concerns global progress, Table 2 shows the most recent generation of evolving agents, up to generation 140,000, that achieved significantly better performance against opponents of the last generation with respect to opponents of previous generations (1-tailed t-tests, N=5 replications). As shown in the table, in the case of the Standard algorithm, coevolution produced global progress up to generation 80,000 and 120,000, in the case of predators and prey respectively. Indeed, predators of generation 80,000 produced significantly better performance against prey of generation 150,000 than predators of previous generations up to generation 30,000. Moreover, prey of generation 120,000 produced significantly better performance against predators of generation 150,000 than prey of previous generations up to generation 70,000. In the case of the Simplified algorithm, global progress continued up to generation 90,000 and 110,000 in the case of predators and prey, respectively. Instead, the Vanilla algorithm did not show any evidence of global progress. Indeed, performance of modern agents were not significantly better than the performance of their ancestors against competitors of generation 150,000.

Overall, these data indicate that the Standard and the Simplified algorithms produced both historical and global progress for prolonged evolutionary periods. The Vanilla algorithm, instead, did not produce neither historical neither general progress.

To verify the relative efficacy of the three methods we post-evaluated predators and prey of generation 150,000 evolved with the standard algorithm against the opponents of the same generation evolved with the Simplified and Vanilla algorithms (Fig. 4). In each comparison, we evaluated the robots evolved in the five replications of the Standard condition against the robots evolved in the five replications of the Simplified and the Vanilla conditions. The predators evolved in the Standard condition achieved significantly higher performance when they were evaluated against the prey evolved in the Simplified ($p < .05$) and Vanilla ($p < .001$) conditions than when they were evaluated against the prey evolved in the Standard condition (Fig. 4, red histograms). Similarly, the prey evolved in the Standard condition achieved significantly higher performance when they were evaluated against the predators evolved in the Simplified ($p < .05$) and the Vanilla ($p < .001$) conditions than when they were evaluated against the predators evolved in the Standard condition (Fig. 4, blue histograms). Therefore, predators and prey evolved in the Standard condition outperformed the agents evolved in the other conditions.

**Table 2. Global progress measured as the most recent generation of evolving agents that achieved significantly better performance against opponents of generation 150,000 than agents of more ancient generations.**

| Agents generation | Standard | | Simplified | | Vanilla | |
|---|---|---|---|---|---|---|
| | Predators | Prey | Predators | Prey | Predators | Prey |
| 1 | 0 | 0 | 0 | 0 | 0 | 0 |
| 2 | 0 | 1 | 0 | 0 | 0 | 0 |
| 3 | 1 | 2 | 2 | 1 | 0 | 0 |
| 4 | 1 | 3 | 2 | 1 | 0 | 0 |
| 5 | 1 | 3 | 2 | 2 | 0 | 0 |
| 6 | 2 | 3 | 3 | 3 | 0 | 0 |
| 7 | 2 | 4 | 4 | 3 | 0 | 0 |
| 8 | 3 | 5 | 4 | 4 | 0 | 1 |
| 9 | 3 | 6 | 5 | 4 | 0 | 0 |
| 10 | 3 | 6 | 5 | 4 | 0 | 0 |
| 11 | 3 | 6 | 5 | 5 | 0 | 0 |
| 12 | 3 | 7 | 5 | 5 | 0 | 0 |
| 13 | 3 | 7 | 5 | 5 | 0 | 0 |
| 14 | 3 | 7 | 5 | 5 | 0 | 0 |

*Note. Results relatively to the three evolving algorithms are separately reported in the left (standard), middle (Simplified), and right (Vanilla) columns. Generations are expressed in x10$_3$, i.e. 1=10000, 2=20000 and so on. Generation 0 refers to the very first generation of randomly generated individuals. Comparison between performances were conducted by means of one-tailed t-test, with N=5 replications.*

Overall, these results indicate that the possibility to filter out opportunistic individuals on the basis of the performance obtained against the validation opponents, which is missing in the Vanilla condition, permit to develop more effective solutions. Moreover, these results indicate that exposing evolving agents to well differentiated opponents and preserving agents capable of defeating strong opponents, which is missing in the Simplified experimental condition, permit to develop better solutions.

The agents evolved displayed quite sophisticated behaviors as can be appreciated from Fig. 5 discussed below, and from the videos available from http://laral.istc.cnr.it/res/predprey2019/. The agents evolved in the Standard experimental condition, in particular, display all the capabilities that we mentioned above: they avoid fixed and moving obstacles; optimize their motion trajectory with respect to multiple constraints; integrate sensory information over time and regulated their behavior accordingly; anticipate the behavior of the competitor; disorient the competitor by producing variable and irregular behaviors and display strategy able to cope with such protean behaviors; adapt on the fly to the behavior of the current competitor. Remarkably, prey display the ability to avoid obstacles and to escape from the predator by moving both in the forward and backward direction, and the ability to alternate phases during which they move in the forward or in the backward direction appropriately, depending on the circumstances.

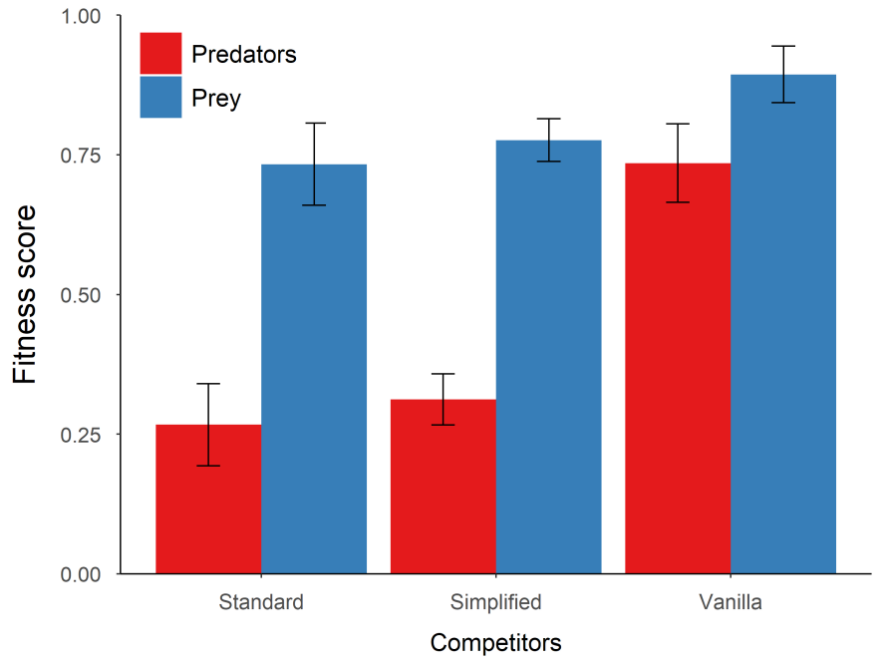

Fig. 4. Cross-experiments test. The histograms display the performance of predators (red) and prey (blue) evolved in the Standard condition evaluated against the opponents evolved in the Standard, Simplified and Vanilla conditions, respectively.

## 4. Behavior complexification

The visual inspection of the robots' behavior indicates that the capabilities of the agents and the articulation of the agents' behavior tend to increase throughout generations. To verify whether and to what extent the behavior of the robots become more complex throughout generations we need to identify a method to measure the complexity of behavior.

Although humans have an intuitive notion of what complexity is, identifying formal way to measure it is far from trivial. One widely used measure of complexity is Shannon's Entropy (Shannon, 1948) that measures the uncertainty of a random variable. This technique can be easily applied to systems composed by multiple discrete components such as cellular automata (Wolfram, 1984). Its application to systems that are continuous and characterized by multi-level and multi-scale organizations, instead, is more challenging. This is the case of the behavior of our robots since it is the result of continuous actions and since it displays a multi-scale organization in which one can identify simple short-term behaviors lasting few hundred milliseconds and long-term behaviors lasting several seconds. An example of short-term behavior is constituted by a phase lasting few hundred milliseconds during which a prey robot turns on the spot to avoid a wall located in its frontal direction. An example of long-term behavior is constituted by a phase during which a predator robot moves along a circular-like trajectory by attempting to reach the prey that moves at full speed along an external circular-like trajectory and by attempting to push the prey against the walls surrounding the arena.

A simple and effective way to measure the complexity of the behavior exhibited by our robots consists in measuring the average derivative of the translational and rotational speed of the robots' wheels calculated on the basis of the following equation:

$$c = \frac{\sum_{t=1}^{s} |\overline{tv}_t - \overline{tv}_{t-1}| + |\overline{rv}_t - \overline{rv}_{t-1}|}{s * 2}$$

where $\overline{tv}$ and $\overline{rv}$ are the desired translational and rotational speeds per seconds, and $s$ is the number of steps.

Fig. 5 shows the complexity level of typical evolved behaviors calculated on the basis of this measure. The Figure has been obtained by: (i) measuring the complexity levels of the behavior exhibited by 80 evolved predators evaluated against 80 evolved prey on the basis of the equation above, (ii) ranking the corresponding 3600 evaluation episodes on the basis of the complexity of the behavior exhibited by predators and prey, and (iii) displaying the behavior of the predators and prey with the simplest, the simple/average, the average, the average/complex, and the more complex behavior (from left to right box, respectively). As can be seen, the measure correctly discriminated between simple behaviors in which the predators move at about constant speed along straight or circular-like trajectories (see left boxes of Fig. 5, both panels) from more complex behaviors in which predators suddenly inverted their direction of motion to block reiterated attempts of the prey to find alternative escaping directions (see far right box of Fig 5, both panels).

This method for measuring complexity is not general. In particular, it can be too simple in the case of agents provided with articulated-limbs in which the variations of the initial joints have much larger effects than the variation of the final joints, or in the case of agents evaluated for significantly long period of time in which the time extension of functionally relevant behaviors can vary widely. However, it is appropriate for the experimental setting considered in this paper. in which the behavior of the agents is regulated by few independent actuators and in which the overall duration of the agents' behavior is limited

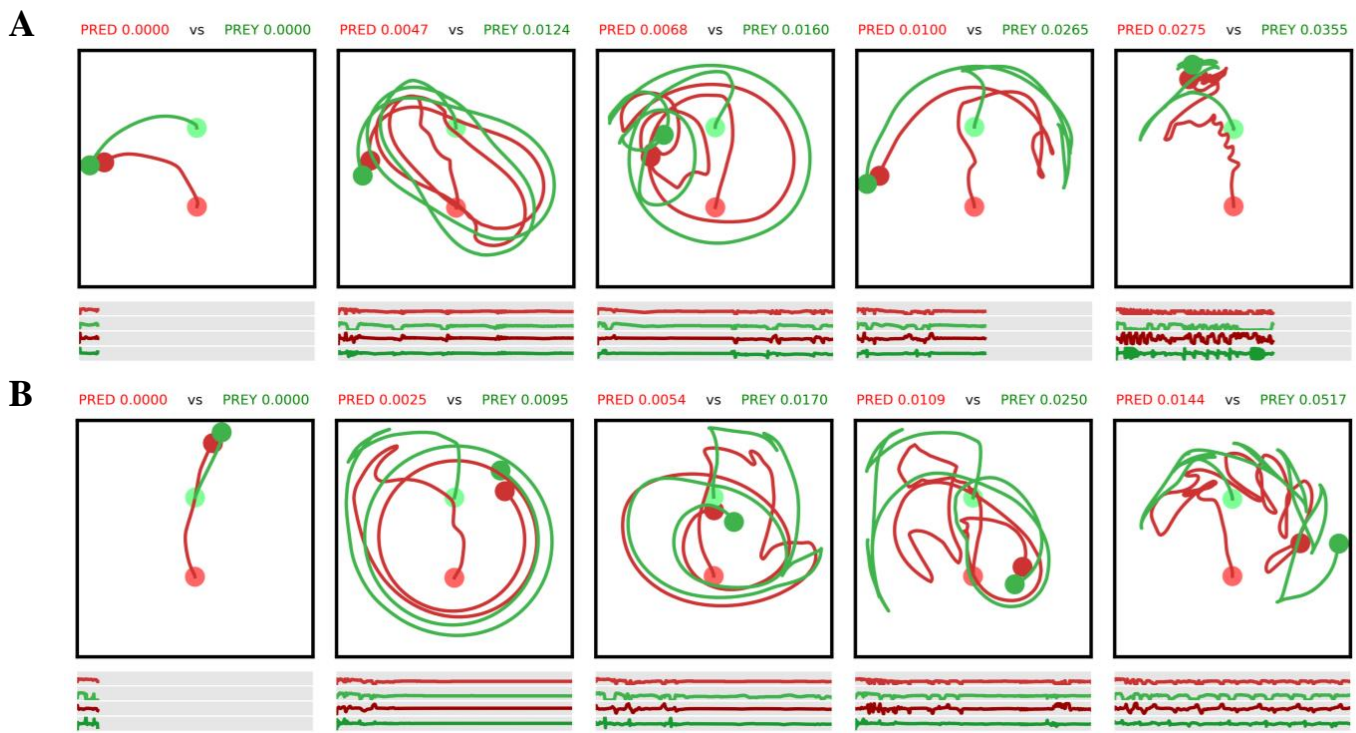

Fig. 5. Representative predators and prey behaviors of agents of generation 150,000 ordered by the behavioral complexity of predators (panel A) or prey (panel B). Each box displays the trajectories of the predator and the prey in red and green. The light and dark circles indicate the starting and final position of the robots in the corresponding evaluation episode. The curves at the bottom of the trajectories indicate the translational and rotational movements of the predator and prey robots over time (red and green lines, respectively). The red and green numbers at the top indicate the behavioral complexity of the behavior of the predators and prey, respectively.

Fig. 6 shows the average complexity level of predator and prey behaviors across generations for the experiments carried with the Standard, Simplified and Vanilla algorithms. The complexity measure is averaged over the individuals of the population obtained in the five replications of the experiment. The behavior of the agents evolved for 150,000 generations in the Standard condition is more complex than the behavior of the agents evolved in the other two conditions (vs Simplified, p<.01; vs Vanilla, p<.01). Moreover, the behaviors of the agents evolved in the Simplified condition is more complex than the behavior of the agents evolved in the Vanilla condition (p<.01).

To verify the duration of the complexification process across generations, we compared the complexity of the behaviors of the robots of each generation (every 10,000 generations) with the complexity of the behaviors of the robots of previous generations. In the case of the Standard condition, the complexity of both predators and prey increased significantly up to generation 90,000 as demonstrated by the fact that the complexity of the behaviors of the robots at this stage is significantly higher than the complexity of the behaviors of the robots in previous generations for both predators and prey. In the case of the experiment performed in the Simplified condition, the complexity of behavior increases up to generation 110,000 for both predators and prey. Instead, in the case of the experiment performed in Vanilla condition, the complexity of behavior does not vary significantly across generations.

The fact that the agents evolved in the Standard condition, which possess the best performing strategies (Fig. 4) display the most complex behavior indicates the presence of a correlation between performance and behavior complexity. This correlation is also supported by the fact that the performance and the complexity of behavior increase across generations until generation 90,000 and 110,000 in the case of the Standard and Simplified experimental conditions.

The fact that the complexity of behavior increases for a greater number of generations in the Simplified condition than in the Standard condition can be explained by considering that, as shown above, the rate of progress of the Simplified condition is lower than the rate of progress of the Standard condition.

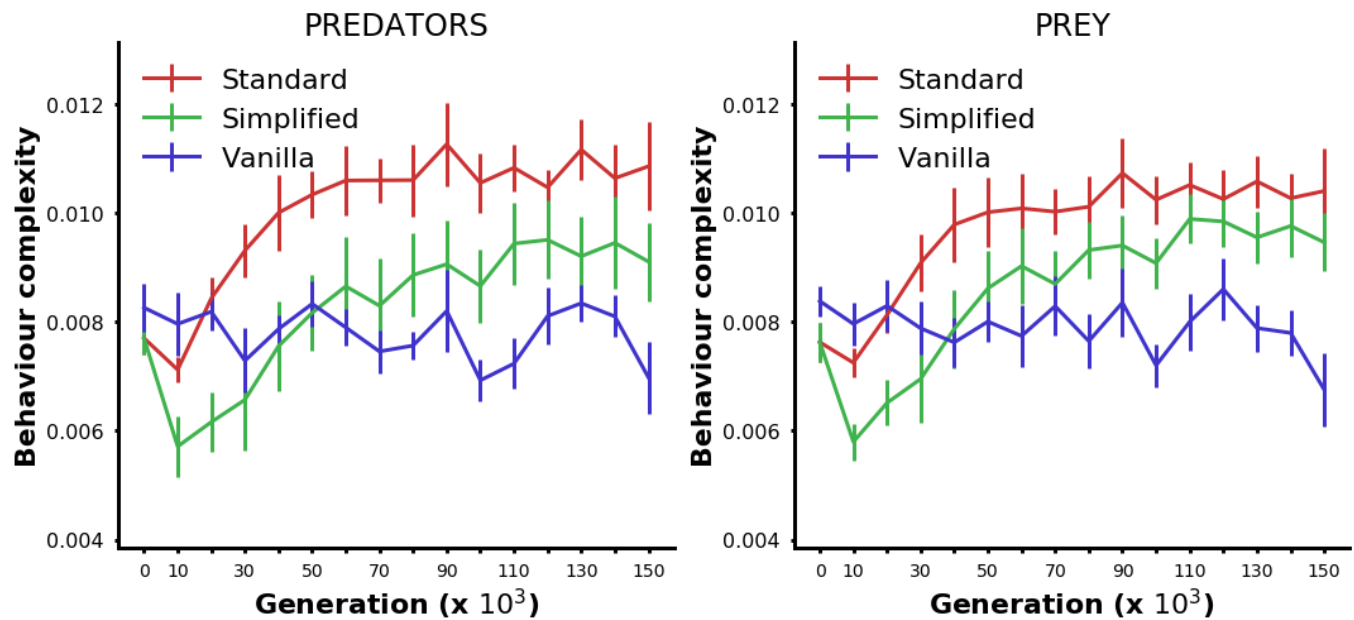

Fig. 6. Behavioral complexity of agents throughout generations in the case of the experiments carried out with the Standard, Simplified, and Vanilla experimental conditions. The left and right picture display the complexity of the behavior of predator and prey robots, respectively. Each point of each curve represents the average complexity of the behavior of all agents of a given generation post-evaluated against all opponents of the same generation, every 10,000 generations. Data averaged over 5 replications. The error bars depict standard errors.

Fig. 7 shows the correlation between the behavioral complexity and the performance of the agents. Performance are measured by post-evaluating the agents against the opponents of the last generation. Performance and behavioral complexity are strongly positively correlated in the case of the Standard and of the Simplified conditions while are not correlated in the case of the Vanilla condition (Fig. 7). Indeed, there is a strong positive correlation in the case of predators (r=.90, p<.01) and prey (r=.95, p<.01) in the Standard condition (Fig.7, left). A similar pattern can be observed in the case of the Simplified experimental condition (Fig. 7, center) for predators (r=.95, p<.01) and prey (r=.96, p<.01). Instead, in the case of the Vanilla condition (Fig.7, right), the two measures are not correlated: predators (r=-,09, p=.75) and prey (r=-.06, p=.83).

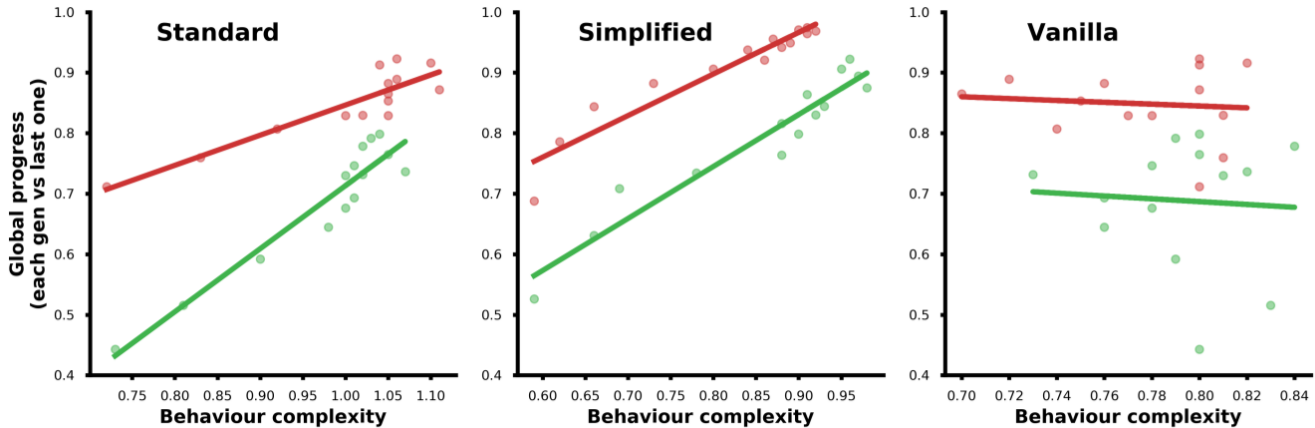

Fig. 7. Correlations between the performance and the behavior complexity of evolving agents. Data for the Standard (left panel), Simplified (central panel), and Vanilla (right panel) conditions. Performance refers to the fitness obtained by post-evaluating the agents against the opponents of the last generation.

Overall, these analyses show that global progress produces to a complexification of the agents' behavior. There is not a one-to-one correspondence between progress in performance and behavior complexification. Progresses in performance arise as a result of variations that increase, maintain, or reduce the complexity of the agents' behavior. The complexification is caused by the fact that the former type of variations is more common than the latter.

## 5. Increasing the complexity of the environment

Previous studies reported evidences indicating that the complexity of the environmental conditions can favor progress (Nolfi and Floreano, 1998: Harrington, Freeman, and Pollack, 2014). To verify whether this can have an impact in the case of the problem studied in this paper, and to verify whether the relative efficacy of the three algorithms compared are influenced by the specificity of the problem considered we ran a new set of experiments in which the environment is more complex. This was realized by adding a cylindrical obstacle with a diameter of 0.1 m in the center of the environment. The obstacle is sufficiently high to visually occlude the opponent. This additional obstacle represents a new opportunity and a new constraint for both species. From the point of view of predators, it constitutes an additional opportunity to push prey in deadlock situations and an additional constraint on the movement of the predators. From the point of view of the prey, it enables to hide from the predators and to shelter from predator attacks but also constitutes a constraint with respect to escape paths.

Fig. 8 shows the master tournament analysis for the three experimental conditions. Overall, the performance of the predators is higher in these experiments than in the basic experiments reported above. However, also in this case none of the two species manage to fully defeat the competing species. Consequently, both species have the possibility to improve during the entire co-evolutionary process.

The dynamics of the co-evolutionary process is qualitatively similar and is characterized by historical and global progress in the Standard and Simplified conditions and by the lack of global progress in the Vanilla condition.

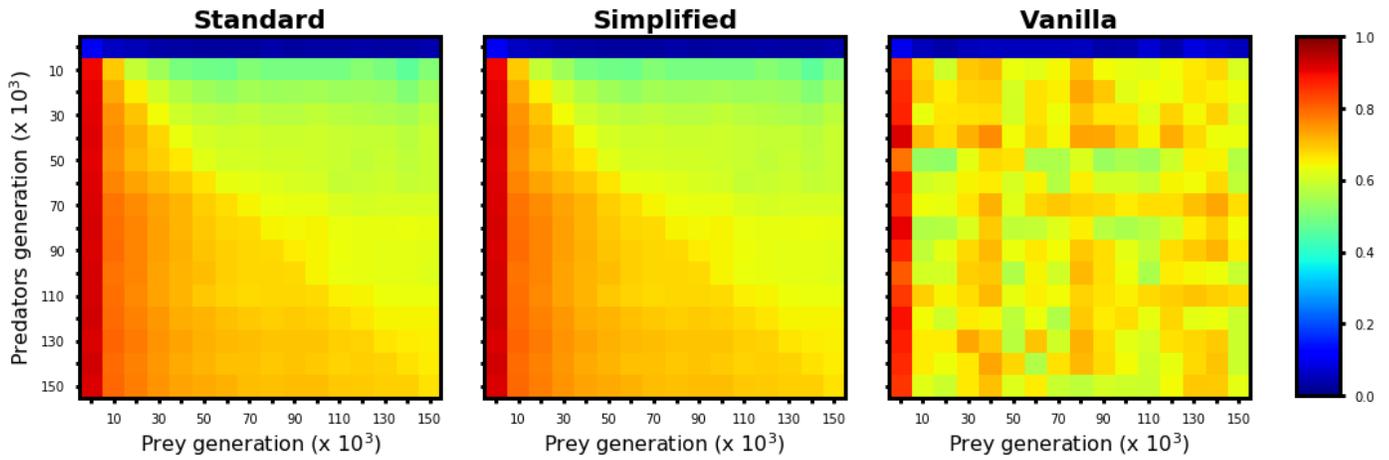

Fig. 8. Performance of predators and prey of every 10,000 generations post-evaluated against competitors of previous and following generations (master tournament), averaged over 5 replications. For each comparison, we indicate the performance of the predator. The performance of the prey corresponds to 1.0 minus the performance of the predators. Results obtained in the Standard (left panel), Simplified (central panel) and Vanilla (right panel) conditions. Each panel shows the average results of 5 replications.

Fig. 9 shows the comparison of the historical and global progress in the basic experiments reported in previous sections and in the new experiments with the additional obstacle. As measure of historical progress, we report the performance of the agents of last generation against the opponents of previous generations (Fig. 9, panel A). As measure of global progress, we report the performance of agents of all generations against the opponents of the last generation (Fig 9. Panel B). Performance have been normalized in the range [0.0, 1.0].

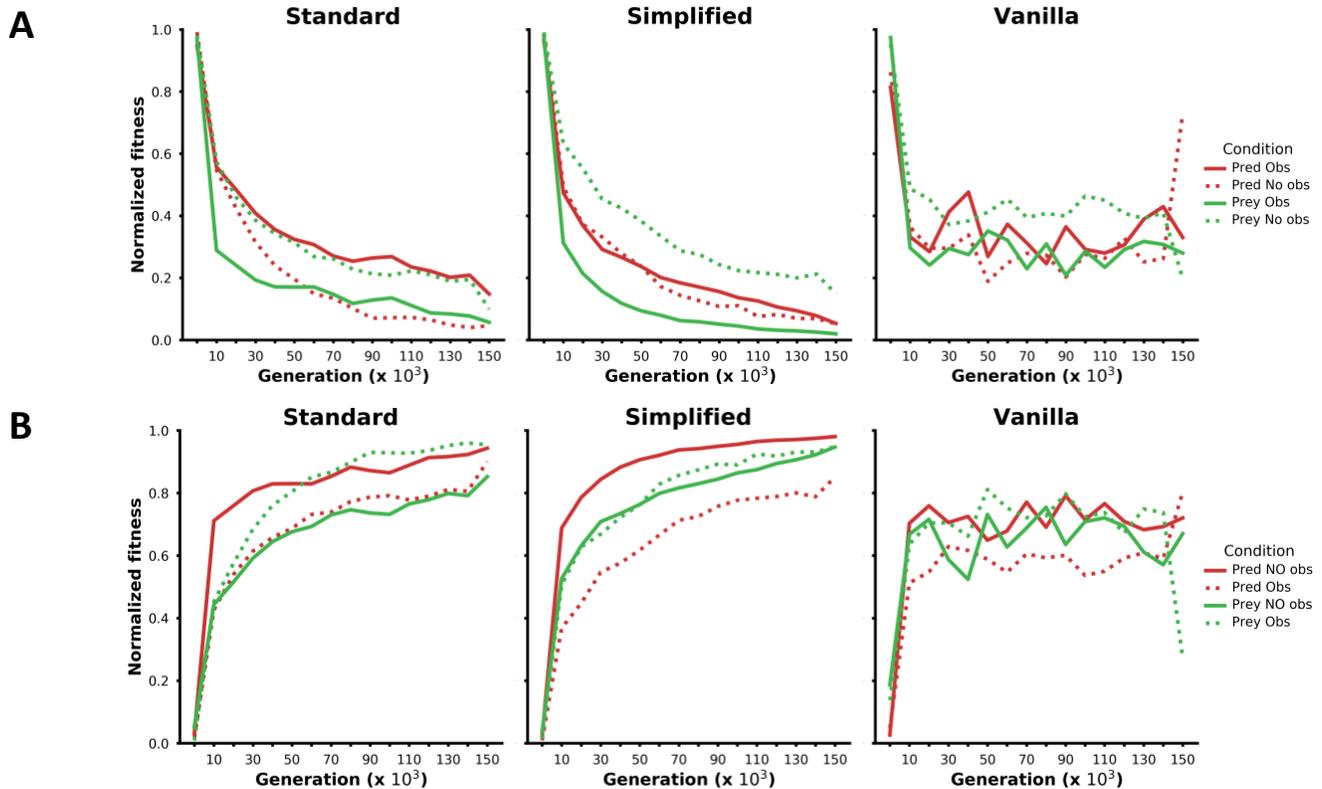

Fig. 9. Panel A displays the performance of the agents of the last generation against opponents of previous generations. Panel B displays the performance of the agents of all generations against agents of the last generation. The left, central,

and right figures show the data obtained in the Standard, Simplified and Vanilla experimental conditions. Performance data have been normalized in the range [0.0, 1.0]. Each figure shows the average result of 5 replications.

As can be seen, the Standard and the Simplified conditions produce consistent historical and global progress during the entire evolutionary process both in the original experiments and in the new experiments with the additional obstacle. The Vanilla condition does not produce historical and global progress after the first 10,000 generations, both in the original experiment and in the new experiment with the additional obstacle. The rate of progress produced by the Standard and Simplified condition in the experiments without and with the additional obstacle is similar.

Consequently, the usage of a more complex environment does not produce an impact on the achievement of historical or global progress.

## 6. Discussion

We demonstrated how a specially designed competitive co-evolutionary method that identify and filter out opportunistic individuals produce long-term historical and global progress. This is achieved by periodically dividing the opponents in two sub-groups of evolving and non-evolving individuals and by using the latter subgroup to validate the generality of the evolving agents. The term historical progress indicates that evolving agents improve their performance against ancient competitors. The term global progress indicates that evolving agents improve their performance against all type of competitors, including competitors that they did not encounter (e.g. future competitors). The term long-term refers to the fact that evolution continues to produce and to accumulate progresses for long evolutionary periods.

As far as we know, this is the first time that competitive co-evolutionary experiments involving embodied and situated agents lead to long-term global process (i.e. global progress over more than 100,000 generations). Indeed, previous related experiments (Cliff and Miller, 1995-1996 Nolfi and Floreano, 1998; Buason, Bergfeldt and Ziemke, 2005; Stanley and Mikkulainen, 2002; Janssen et al. 2016) did not show evidences of global progress and were carried on for only few hundreds of generations.

From a theoretical point of view, the selection of agents that generalize with respect to a limited number of validation opponents does not guarantee global progress. On the other hand, as pointed out by Wagner (2011), the numbers of solutions that are effective in variable environmental conditions tend to decrease exponentially with the increase of the number of environmental conditions. This implies that the solutions that are effective against a sufficiently large number of opponents can be general, i.e. can generalize to a wide range of opponents.

Finally, we showed how exposing agents to well differentiated opponents and preserving agents capable of defeating strong opponents increase the rate of progress and lead to better solutions.

The method proposed present elements introduced in other algorithms described in the literature but combine them in new ways. In particular, the usage of a subset of the opponents for validation presents similarities with techniques used in neural networks to reduce/eliminate overfitting (Searle, 1995; Srivastava et al., 2014). The selection of well-differentiated opponents and the preservation of agents capable of defeating strong opponents have similarities with multi-objective optimization methods (Debb, 2001; Coello et al., 2007) and with technique used to preserve population diversity (Laumanns et al., 2002; Mouret and Doncieux, 2009).

The accumulation of global progress over many generations leads to high effective solutions that involve the production of rather articulated behaviors. More specifically, although there is not a one to one correspondence between progress and behavior complexification, the behavior of the evolving agents tends to become more complex across generations.

**Appendix**

**A1. The robots and the environment**

The robots were simulated MarxBot (Bonani et al., 2010), i.e. circular robots with a diameter of 17cm equipped with a differential drive motion system, a ring of 24 color LEDs, 24 infrared sensors, 4 ground sensors, an omnidirectional camera, and a traction sensor. Predator and prey robots had their LEDs turned on in red and green respectively. The robots were situated in a 3x3 m square arena surrounded by black walls. The ground was colored in grayscale with a level of darkness that varies linearly from full white to full black from the center to the periphery of the arena (see Fig. A1, left panel).

The maximum speed (*ms*) that the wheels of the differential drive motion system could assume was 10 and 8.5 rad/s, for the prey and the predators respectively, in the case of the experiments reported in Section 3 and 4, and 10 rad/s, for both the prey and the predators, in the case of the experiment reported in Section 5. The relative speed of the two robots was tuned to balance approximately the overall complexity of problem faced by predators and prey, i.e. to avoid that the average fitness of one species reach the maximum or the minimum value.

The experiments were run in simulation by using the FARSA open-software tool that includes an accurate simulator of the robots and of the environment (Massera, Ferrauto, Gigliotta and Nolfi, 2014). FARSA has been used to successfully transfer results obtained in simulation to hardware in similar experimental settings (Baldassarre et al., 2007; Sperati et a., 2008). The update frequency of state of the environment, of the robots, of the robots' sensors and motors, and of the robots' neural network was10Hz.

The experiments reported in this paper can be replicated by downloading and installing FARSA, which is available from https://sourceforge.net/projects/farsa/, and by downloading and installing the experimental plugin available from http://laral.istc.cnr.it/res/predprey2019/ Simione_Nolfi_2019_plugin.zip. Videos of the evolved behaviors are available from http://laral.istc.cnr.it/res/predprey2019/.

**A2. The neural network controller**

Each robot is provided with a neural network controller that includes 25 sensory neurons, 10 internal neurons with recurrent connections, and 2 motor neurons (Fig. A1, right panel).

The sensory layer includes 8 sensory neurons that encode the average activation state of eight groups of three adjacent infrared sensors, 8 neurons that encode the fraction of green or red light perceived in the eight 45° sectors of the visual field of the camera, 1 neuron that encodes the average amount of green or red light detected in the entire visual field of the camera, 4 neuron that encode the state of the four ground sensors, 1 neuron that encodes the average activation of the four ground sensor, 1 neuron that encodes whether the robot collides with an obstacle (i.e. whether the traction force detected by the traction sensor exceed a threshold), 1 clock neuron that encodes the time passed since the beginning of the trial, and 1 simulated fatigue neuron that encodes the "tiredness" of the robot, (i.e. the amount of energy recently spent by the robot, see eq. 1). The state of the sensory neurons was normalized in the range [0.0, 1.0].

The motor layer included two motor neurons (*tm* and *rm*) that encoded the desired translational and rotational motion of the robot in the range [0.0, 1.0].

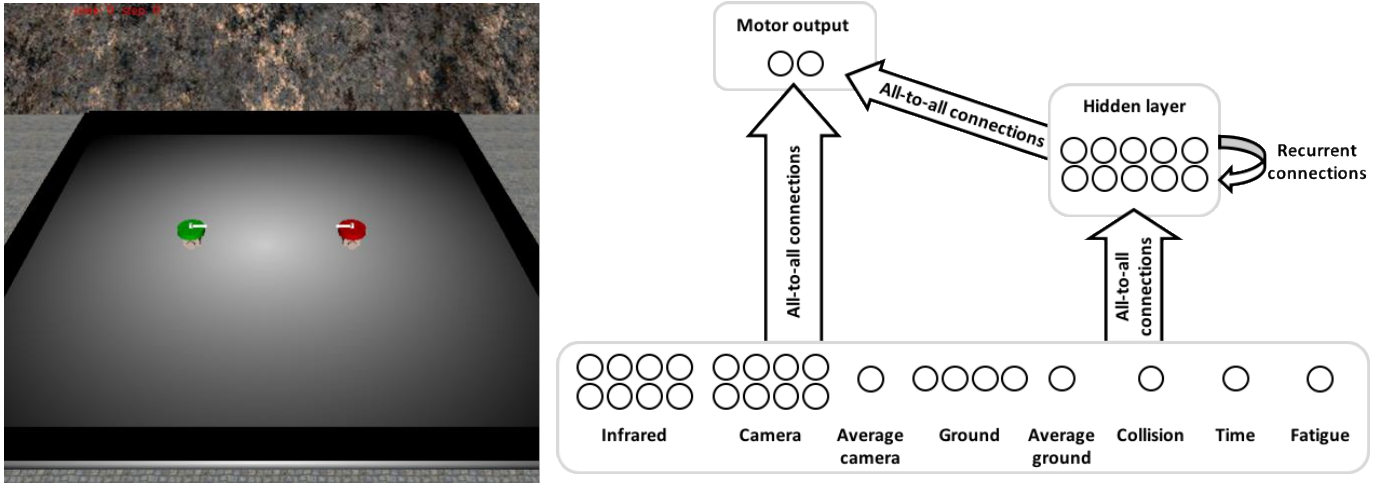

Fig. A1. Left panel: the robots and the environment in simulation. The red and green robots correspond to the predator and prey robot, respectively. Right panel: the neural network controller. Black arrows indicate all-to-all feedforward connections between two layers. All-to-all recurrent connections are present between the hidden neurons.

The max speed at each time step ($ms_t$) is modulated on the basis of the tiredness of the robot at time $t$ ($tir_t$), i.e. the energy spent by the robot to move during the last 20 s. They are calculated on the basis of the following equations:

$$tir_t = \left( \sum_{t=-200}^{0} rs_t / 200 \right)^2 \quad (1)$$

$$ms_t = ms(1 - tir_t) \quad (2)$$

where *ms* was the maximum speed of the wheels described above and $rs_t$ are the absolute speed of the two wheels at time $t$ normalized in the range [0.0, 1.0].

The desired rotational speed of the left and right wheels (*rsl* and *rsr*) at time *t* are calculated on the basis of following equations:

$$rsl_t = \begin{cases} ms_t \times rm_t \times f(tm_t) & if\, tm_t < 0.5 \\ ms_t \times rm_t & otherwise \end{cases} \quad (3)$$

$$rsr_t = \begin{cases} ms_t \times rm_t \times f(tm_t) & if\, tm_t > 0.5 \\ ms_t \times rm_t & otherwise \end{cases} \quad (4)$$

$$f(x) = -2^3(x - 0.5)^2 + 1 \quad (5)$$

The dependency of the maximum speed at which predators and prey can move on the tiredness (i.e. on the amount of energy recently spent by the robot) encourages evolving robots to avoid wasting energy.

### A3. Evolving parameters and fitness calculation

The connection weights and the biases of the neural network controller of each robot are encoded in a vector of 442 floating-point values (genotype). The genotypes of the two populations at generation 0 is filled with values generated with a random uniform distribution in the range [-5.0, 5.0].

Predators were evolved for the ability to capture prey (i.e. to reach and physically touch the prey) as fast as possible and prey were evolved for the ability to avoid being captured as long as possible. Each robot is evaluated against *n* competitors, one competitor at a time, during *n* corresponding trials. At the beginning of each trial the predator and prey robots are placed on the central-left and central-right side of the environment facing toward each other and are allowed to move for 1000 steps that correspond to 100 s (Fig A1, left). The fitness of the predator is the fraction of time required by the predator to capture the prey. The fitness of the prey is the inverse of the fraction of time required by the predator to capture the prey, respectively. The total fitness is the average fitness obtained during the episodes in which the agents is evaluated against different opponents.